\documentclass[conference]{IEEEtran}
\usepackage[latin9]{inputenc}
\usepackage{textcomp}
\usepackage{amsmath}
\usepackage{graphicx}
\usepackage[unicode=true]
 {hyperref}

\makeatletter
\ifCLASSINFOpdf
\else
\fi
\@ifundefined{showcaptionsetup}{}{%
 \PassOptionsToPackage{caption=false}{subfig}}
\usepackage{subfig}
\makeatother

\begin{document}

\title{Integer Factorization with a Neuromorphic Sieve}

\author{\IEEEauthorblockN{John~V.~Monaco~and~Manuel~M.~Vindiola} \IEEEauthorblockA{U.S.~Army Research Laboratory\\
Aberdeen Proving Ground, MD 21005\\
Email: john.v.monaco2.civ@mail.mil, manuel.m.vindiola.civ@mail.mil}}

\maketitle

\begin{abstract}
The bound to factor large integers is dominated by the computational
effort to discover numbers that are smooth, typically performed by
sieving a polynomial sequence. On a von Neumann architecture, sieving
has log-log amortized time complexity to check each value for smoothness.
This work presents a neuromorphic sieve that achieves a constant time
check for smoothness by exploiting two characteristic properties of
neuromorphic architectures: constant time synaptic integration and
massively parallel computation. The approach is validated by modifying
msieve, one of the fastest publicly available integer factorization
implementations, to use the IBM Neurosynaptic System (NS1e) as a coprocessor
for the sieving stage.
\end{abstract}

\IEEEpeerreviewmaketitle{}

\section{Introduction}

A number is said to be \emph{smooth} if it is an integer composed
entirely of small prime factors. Smooth numbers play a critical role
in many interesting number theoretic and cryptography problems, such
as integer factorization \cite{pomerance1994role}. The presumed difficulty
of factoring large composite integers relies on the difficultly of
discovering many smooth numbers in a polynomial sequence, typically
performed through a process called \emph{sieving}. The detection and
generation of smooth numbers remains an ongoing multidisciplinary
area of research which has seen both algorithmic and implementation
advances in recent years \cite{granville2008smooth}.

This work demonstrates how current and near future neuromorphic architectures
can be used to efficiently detect smooth numbers in a polynomial sequence.
The \emph{neuromorphic sieve} exploits two characteristic properties
of neuromorphic architectures to achieve asymptotically lower bounds
on smooth number detection, namely constant time synaptic integration
and massively parallel computation. Sieving is performed by a population
of leaky integrate-and-fire (LIF) neurons whose dynamics are simple
enough to be implemented on a range of current and future architectures.
Unlike the traditional CPU-based sieve, the factor base is represented
in space (as spiking neurons) and the sieving interval in time (as
successive time steps). Integer factorization is achieved using a
neuromorphic coprocessor for the sieving stage, alongside a CPU.

\section{Integer Factorization \label{sec:Integer-factorization}}

Integer factorization is presumed to be a difficult task when the
number to be factored is the product of two large primes. Such a number
$n=pq$ is said to be a \emph{semiprime}\footnote{Not to be confused with \emph{pseudoprime}, which is a probable prime.}
for primes $p$ and $q$, $p\ne q$. As $\log_{2}n$ grows, i.e.,
the number of bits to represent $n$, the computational effort to
factor $n$ by trial division grows exponentially.

Dixon's factorization method attempts to construct a congruence of
squares, $x^{2}\equiv y^{2}\mod n$. If such a congruence is found,
and $x\not\equiv\pm y\mod n$, then $\gcd\left(x-y,n\right)$ must
be a nontrivial factor of $n$. A class of subexponential factoring
algorithms, including the quadratic sieve, build on Dixon's method
by specifying how to construct the congruence of squares through a
linear combination of smooth numbers \cite{pomerance1984quadratic}.

Given smoothness bound $B$, a number is $B$-smooth if it does not
contain any prime factors greater than $B$. Additionally, let ${\bf v}=\left[e_{1},e_{2},\dots,e_{\pi\left(B\right)}\right]$
be the exponents vector of a smooth number $s$, where $s=\prod_{1\le i\le\pi\left(B\right)}p_{i}^{v_{i}}$
, $p_{i}$ is the $i\mbox{th}$ prime, and $\pi\left(B\right)$ is
the number of primes not greater than $B$. With a set of $\pi\left(B\right)+1$
unique smooth numbers $S=\left\{ s_{1},s_{2},\dots,s_{\pi\left(B\right)+1}\right\} $,
a perfect square can be formed through some linear combination of
the elements of $S$, $y^{2}=\prod_{s_{i}\in S}s_{i}$. The reason
for this is that there exists at least one linear dependency among
a subset of the $\pi\left(B\right)+1$ exponents vectors that contain
$\pi\left(B\right)$ elements each. Gaussian elimination or block
Lanczos algorithm can be used to uncover this linear dependency \cite{montgomery1995block}.

Smooth numbers are detected by sieving a polynomial sequence. Sieving
relies on the fact that for each prime $p$, if $p\mid f\left(x\right)$
then $p\mid f\left(x+ip\right)$ for any integer $i$ and polynomial
$f$. To sieve the values $f\left(x\right)$, for $0\le x<M$, on
a von Neumann architecture, a length $M$ array is initialized to
all zeros. For each polynomial root $r$ of each prime $p$ in the
factor base, $\ln p$ is added to array locations $r+ip$ for $i=0,1,\dots$,$\frac{M}{p}$.
This step can be performed using low precision arithmetic, such as
with integer approximations to $\ln p$. After looping through each
prime in the factor base, array values above a certain threshold will
correspond to polynomial values that are smooth with high probability\footnote{By exploiting the fact that $\ln ab=\ln a+\ln b$.}.
This process is referred to hereafter as CPU-based sieving.

Due to errors resulting from low precision arithmetic, some of the
array values above the threshold will end up being not smooth and
some values that are smooth will remain below the threshold. The threshold
controls a tradeoff between the false positive rate (FPR) and false
negative rate (FNR), from which a receiver operating characteristic
(ROC) curve is obtained. Since the actual factorizations are lost
after sieving, the smooth candidates must subsequently be factored
over $\mathcal{F}$, which also serves as a definite check for smoothness.
Factoring a number with small prime factors can be done efficiently,
and this effort can be neglected as long as there are not too many
false positives \cite{crandall2006prime}.

The quadratic sieve \cite{pomerance1984quadratic} detects smooth
numbers of the form 
\begin{equation}
f\left(x\right)=\left(x+\lceil\sqrt{n}\rceil\right)^{2}-n\label{eq:quadratic-sieve}
\end{equation}
where $x=-\frac{M}{2},\dots,\frac{M}{2}-1$. The factor base $\mathcal{F}$
contains primes $p$ up to $B$ such that $n$ is a quadratic residue
modulo $p$, i.e., $r^{2}\equiv n\mod p$ for some integer $r$. This
ensures that each prime in the factor base (with the exception of
2) has two modular roots to the equation $f\left(x\right)\equiv0\mod p$,
increasing the probability that $p\mid f\left(x\right)$. If $\mathcal{F}$
contains $b$ primes, then at least $b+1$ smooth numbers are needed
to form the congruence of squares.

It is the sieving stage of the quadratic sieve that is the focus of
this work. Sieving comprises the bulk of the computational effort
in the quadratic sieve and the relatively more complex number field
sieve (NFS) \cite{crandall2006prime}. On a von Neumann architecture,
sieving requires at least $\frac{1}{2}M+M\sum_{p\in\mathcal{F}\setminus2}\frac{2}{p}$
memory updates where $\mathcal{F}\setminus2$ is the set of factor
base primes excluding 2. Given likelihood $u^{-u}$ of any single
polynomial value being smooth, where $u=\frac{1}{2}\frac{\ln n}{\ln B}$,
an optimal choice of $B$ is $\exp\left(\frac{1}{2}\sqrt{\ln n\ln\ln n}\right)$
\cite{crandall2006prime}. This yields a total runtime of $B^{2}$,
where the amortized time to sieve each value in the interval is $\ln\ln B$.

\section{Neuromorphic Sieve \label{sec:Neuromorphic-sieve}}

The neuromorphic sieve represents the factor base in space, as tonic
spiking neurons, and the sieving interval in time through a one-to-one
correspondence between time and the polynomial sequence. Let $t\equiv\left(x-x_{min}\right)$,
where $t$ is time, $x$ is the sieving location that corresponds
to $t$, and $x_{min}$ is the first sieving value. Then polynomial
values can be calculated by $f\left(x\right)=f\left(t+x_{min}\right)$.
Each tonic neuron corresponds to a prime (or a power of a prime) in
the factor base and spikes only when it divides the current polynomial
value. If enough neurons spike at time $t$, then $f\left(x\right)$
is likely smooth since each neuron represents a factor of $f\left(x\right)$.
This formulation reverses the roles of space and time from the CPU-based
sieve, in which the sieving interval is represented in space, as an
array, and a doubly nested loop iterates over over primes in the factor
base and locations in the sieving interval.

Construction of the neuromorphic sieve is demonstrated through an
example using the semiprime $n=91$, quadratic polynomial $f(x)=\left(x+\lceil\sqrt{91}\rceil\right)^{2}-91$,
and sieving interval $x=-5,\dots,4$. The smoothness bound is set
to $B=5$. This creates a prime factor base $\mathcal{F}=\{2,3,5\}$,
the primes up to $B$ such that the Legendre symbol $\left(\frac{n}{p}\right)=1$,
i.e., $n$ is a quadratic residue modulo each of the primes in the
factor base. Sieving is also performed with prime powers, $p^{e}$
for $e>1$, that do not exceed the magnitude of the polynomial, namely
$3^{2}$, $3^{3}$, and $5^{2}$. Powers of 2 are not included since
they do not have any modular roots to the equation $f\left(x\right)\equiv0\mod2^{e}$
for $e>1$.

\begin{figure}
\begin{centering}
\includegraphics[width=1\columnwidth]{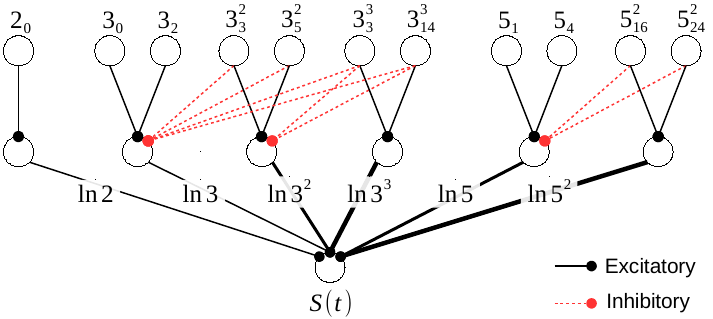}
\par\end{centering}
\caption{Example neuromorphic sieve network. Top layer contains tonic spiking
neurons; middle layer selects the highest prime power; bottom layer
performs the test for smoothness. \label{fig:Neuromorphic-sieve}}
\end{figure}

To achieve a constant-time check for smoothness, a spiking neural
network is composed of three layers that form a tree structure (Figure
\ref{fig:Neuromorphic-sieve}). The top layer contains tonic spiking
neurons for each prime in the factor base as well as prime powers,
up to the magnitude of the polynomial. The middle layer selects the
highest prime power that divides the polynomial value. The bottom
layer contains a single neuron that performs a test for smoothness
by integrating the log-weighted factors. The middle and bottom layers
are stateless, while the dynamics in the top layer encode successive
polynomial values.

For each modular root $r$ of each prime power $p^{e}$ (including
factor base primes, for which $e=1$), designate a neuron that will
spike with period $p^{e}$ and phase $r$ (Figure \ref{fig:Neuromorphic-sieve},
top layer, given by $p_{r}^{e}$, where subscripts denote the modular
root). Due to the equivalence between $t$ and $f\left(x\right)$,
this neuron will spike only when $p^{e}\vert f\left(x\right)$. It
is also the case that if $p^{e}\vert f\left(x\right)$ then $p^{e}\vert f\left(x+ip^{e}\right)$
for any integer $i$, thus only tonic spiking behavior for each modular
root is required.

The tonic spiking neurons are connected through excitatory synapses
to a postsynaptic neuron that spikes if either modular root spikes,
i.e., if $p^{e}\vert f\left(x\right)$ for modular roots $r_{1}$
and $r_{2}$ of prime $p^{e}$ (Figure \ref{fig:Neuromorphic-sieve},
middle layer). Using a LIF neuron model, this behavior is achieved
using synapse weights of the same magnitude as the postsynaptic neuron
threshold. Inhibitory connections are formed between the prime powers
in the top layer and lower prime powers in the middle layer so that
higher powers suppress the spiking activity of the lower powers. This
ensures that only the highest prime power that divides the current
polynomial value is integrated by the smoothness neuron (Figure \ref{fig:Neuromorphic-sieve},
bottom layer). Synapse connections to the smoothness neuron are weighted
proportional to $\ln p^{e}$. The smoothness neuron compares the logarithmic
sum of factors to threshold $\ln f\left(x\right)$ to determine whether
$f\left(x\right)$ can be completely factored over the neurons that
spiked.

\begin{figure}
\begin{centering}
\includegraphics[width=0.96\columnwidth]{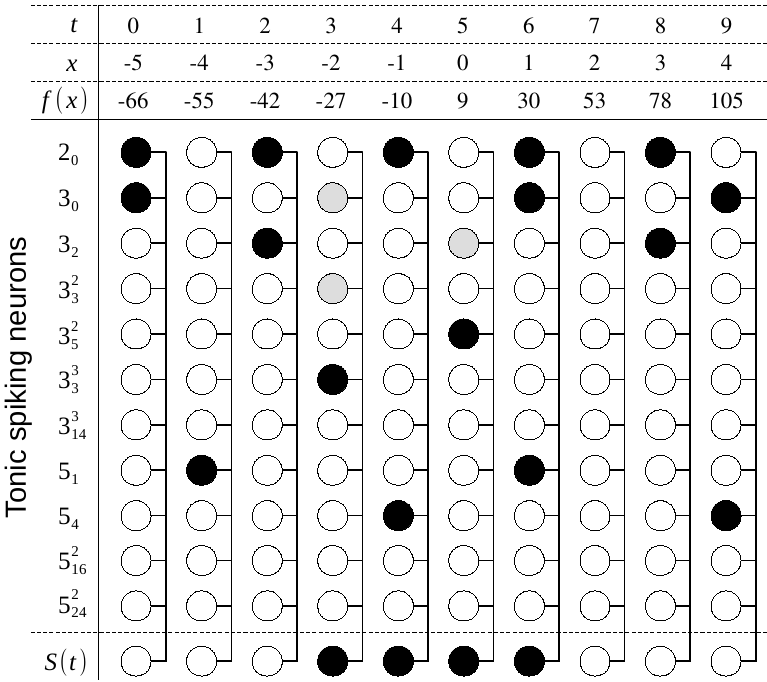}
\par\end{centering}
\caption{Neuromorphic sieve example. The smoothness neuron $S$ spikes when
smooth values are detected. Active neurons on each time step are shown
in black and tonic neurons suppressed by higher prime powers are shown
in gray.\label{fig:Neuromorphic-sieve-example}}
\end{figure}

Figure \ref{fig:Neuromorphic-sieve-example} depicts neuron activity
of the top and bottom layers over time. Activity from tonic neurons
$3_{0}$ and $3_{3}^{2}$ is suppressed by neuron $3_{3}^{3}$ at
time $t=3$, which ensures only $\ln3^{3}$ is integrated by the smoothness
neuron. A similar situation occurs at time $t=5$. The smoothness
neuron spikes at times $\{3,4,5,6\}$ to indicate that polynomial
values $\{-27,-10,9,30\}$ are detected as smooth\footnote{The $-1$ is treated as an additional factor and is easily accounted
for during the linear algebra stage \cite{crandall2006prime}.}.

\section{TrueNorth Implementation \label{sec:TrueNorth-implementation}}

The TrueNorth architecture employs a digital LIF neuron model given
by 
\begin{equation}
V\left(t\right)=V\left(t-1\right)+\sum_{i}A_{i}\left(t-1\right)w_{i}+\lambda\label{eq:lif-neuron}
\end{equation}
where $V\left(t\right)$ is the membrane potential at time $t$, $A_{i}\left(t\right)$
and $w_{i}$ are the $i\mbox{th}$ spiking neuron input and synaptic
weight, respectively, and $\lambda$ is the leak \cite{merolla2014million}.
The neuron spikes when the membrane potential crosses a threshold,
$V\left(t\right)\ge\alpha$, after which it resets to a specified
reset membrane potential, $R$. Each neuron can also be configured
with an initial membrane potential, $V_{0}$. This behavior is achieved
using a subset of the available parameters on the TrueNorth architecture
\cite{merolla2014million}.

For each polynomial root $r$ of each prime power $p^{e}$ (including
factor base primes, i.e., $e=1$), a tonic spiking neuron is configured
with period $p^{e}$ and phase $r$, spiking only when $p^{e}\mid f\left(t+x_{min}\right)$.
This is accomplished by setting $\alpha=p^{e}$, $V_{0}=-\left(r+1-x_{min}\right)\mod p^{e}$,
and $\lambda=1$. Tonic spiking neurons are configured for prime powers
up to $2^{18}$, the maximum period that can be achieved by a single
TrueNorth neuron with non-negative membrane potential ($\alpha$ is
an unsigned 18-bit integer). Tonic neurons are connected to the middle
layer (factor neurons) through excitatory and inhibitory synapses
with weights to invoke or suppress a spike.

TrueNorth implements low-precision synaptic weights through shared
axon types. Each neuron can receive inputs from up to 256 axons, and
each axon can be assigned one of four types. For each neuron, axons
of the same type share a 9-bit signed weight. Thus, there can be at
most 4 unique weights to any single neuron, and all 256 neurons on
the same core must share the same permutation of axon types. This
constraint requires the smoothness neuron weights to be quantized.

Four different weight quantization strategies are evaluated: \emph{regress}
fits a single variable regression tree, i.e., step function, with
4 leaf nodes, to the log factors; \emph{inverse} fits a similar regression
tree using a mean squared error objective function with factors weighted
by their log inverse. This forces quantized weights of smaller, frequent
factors to be more accurate than large factors; \emph{uniform} assigns
each factor a weight of 1, thus the smoothness neuron simply counts
the number of factors that divide any sieve value; \emph{integer}
assigns each factor a weight equal to the log factor rounded to the
nearest integer.

The four quantization strategies are summarized in Figure \ref{fig:Weight-quantization-methods.}.
Using only integer arithmetic, the integer strategy is optimal as
it most closely approximates the log function. The uniform strategy
is a worst case in which only binary (0 or 1) weights are available.
Note that only the regression, inverse, and uniform strategies, which
have at most 4 unique weights, are able to run on TrueNorth. The integer
strategy exceeds the limit of 4 unique weights to any single neuron,
thus is not compatible with the architecture.

A single neuron $S$ performs the smoothness test by integrating the
postsynaptic spikes of the weighted factor neurons. The smoothness
neuron is stateless and spikes only when tonic spiking neurons with
sufficient postsynaptic potential have spiked on each time step. The
stateless behavior of the smoothness neuron is achieved by setting
$\alpha=0$, $R=0$, membrane potential floor to $0$, reset behavior
$\kappa=1$, and $\lambda=\tau$, where $\tau$ is a smoothness threshold.
Postsynaptic spiking behavior is given by 
\begin{equation}
S\left(t+1\right)=\begin{cases}
1 & \mbox{if }\sum A_{i}\left(t\right)w_{i}\ge\tau\\
0 & \mbox{otherwise}
\end{cases}\label{eq:smoothness-neuron}
\end{equation}
where $w_{i}$ and $A_{i}\left(t\right)$ are the weight and postsynaptic
spiking activity of the $i\mbox{th}$ factor neuron, respectively.
TrueNorth is a pipelined architecture, and the postsynaptic spikes
from the tonic spiking neurons at time $t$ are received by the smoothness
neuron at time $t+2$. As a result, the smoothness neuron spikes when
$f\left(t+x_{min}-2\right)$ is likely smooth.

\section{Results \label{sec:Experimental-results}}

We modified \texttt{msieve}\footnote{msieve version 1.52, available at \href{https://sourceforge.net/projects/msieve/}{https://sourceforge.net/projects/msieve/}. }
to use the IBM Neurosynaptic System (NS1e) \cite{merolla2014million}
for the sieving stage of integer factorization. The NS1e is a single-chip
neuromorphic system with 4096 cores and 256 LIF neurons per core,
able to simulate a million spiking neurons while consuming under 100
milliwatts at a normal operating frequency of 1 KHz. \texttt{msieve}
is a highly optimized publicly available implementation of the multiple
polynomial quadratic sieve (MPQS, a variant of the quadratic sieve)
and NFS factoring algorithms. The core sieving procedure is optimized
to minimize RAM access and arithmetic operations. Sieving intervals
are broken up into blocks that fit into L1 processor cache. The host
CPU used in this work is a 2.6GHz Intel Sandy bridge, which has an
L1 cache size of 64KB. Thus, a sieving interval of length $M=2^{17}$
would be broken up into two blocks that each fit into L1 cache.

Results are obtained for $n$ ranging from 32 to 64 bits in increments
of 2, with 100 randomly-chosen $p$ and $q$ of equal magnitude in
each setting for a total of 1700 integer factorizations. This range
was chosen to limit the number of connections to the smoothness neuron
to below 256, as topological constraints are not addressed in this
work. For each $n$, $B$ is set to $\exp\left(\frac{1}{2}\sqrt{\ln n\ln\ln n}\right)$.
The factor base size $b$ ranges from 18 for 32-bit $n$ to 119 for
64-bit $n$. Including prime powers, this requires 93 tonic spiking
neurons for 32-bit $n$ and 429 tonic spiking neurons for 64-bit $n$,
having 47 and 215 connections to the smoothness neuron, respectively.
The sieving interval $M$ is set to $2^{17}$, large enough to detect
$b+1$ $B$-smooth numbers in all but 5 cases in which a sieving interval
of length $2^{18}$ is used.

The factor base primes for each $n$ are determined by \texttt{msieve}
and then used to construct the neuromorphic sieve on the TrueNorth
architecture, as described in Section \ref{sec:TrueNorth-implementation}.
Polynomial roots of the prime factors are calculated efficiently by
the Tonelli-Shanks algorithm, and roots of prime powers are obtained
through Hensel lifting. The resulting network is deployed to the NS1e,
which runs for $M+2$ time steps. On each time step, if a smooth value
is detected, a spike is generated and transmitted to the host which
then checks the corresponding polynomial value for smoothness.

\begin{figure}
\begin{centering}
\subfloat[Weight quantization.\label{fig:Weight-quantization-methods.}]{\begin{centering}
\includegraphics[width=0.47\columnwidth]{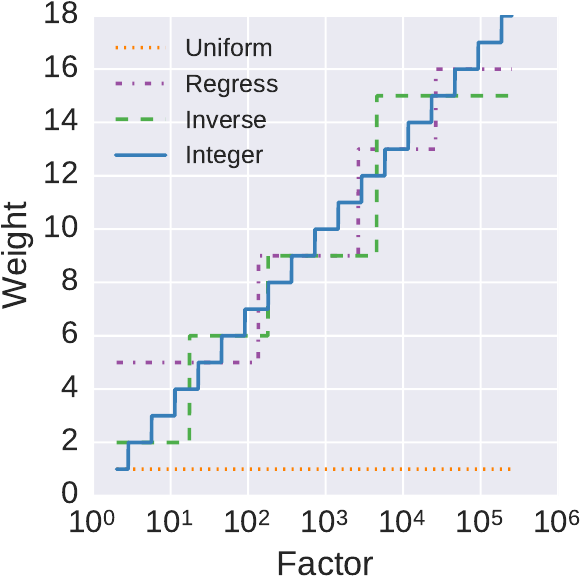}
\par\end{centering}
}\subfloat[64-bit $n$ ROC curves.\label{fig:ROC-curve-summary}]{\begin{centering}
\includegraphics[width=0.48\columnwidth]{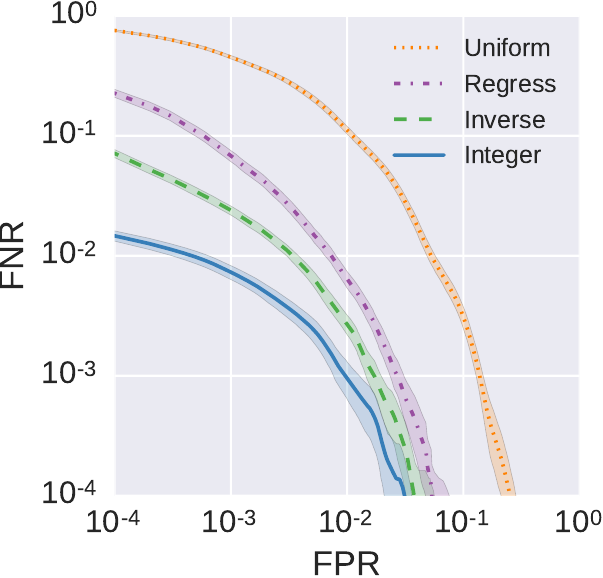}
\par\end{centering}
}
\par\end{centering}
\begin{centering}
\subfloat[Clock cycles/sieve value vs $n$ bits.\label{fig:Ticks-vs-bits}]{\begin{centering}
\includegraphics[width=0.47\columnwidth]{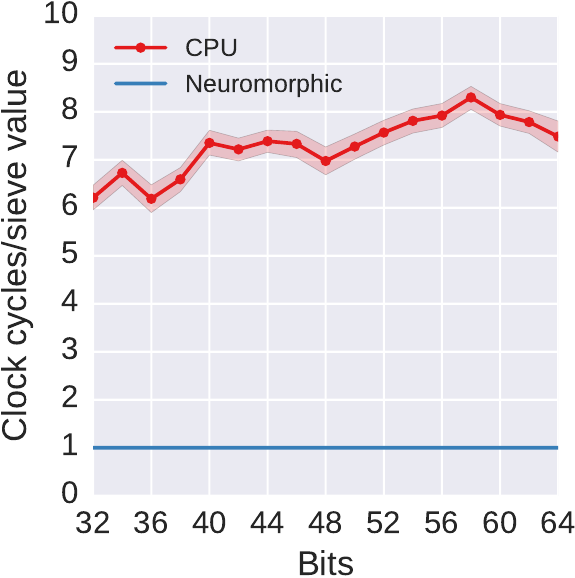}
\par\end{centering}
}\subfloat[FPR given CPU TPR vs $n$ bits.\label{fig:FPR-vs-bits}]{\begin{centering}
\includegraphics[width=0.48\columnwidth]{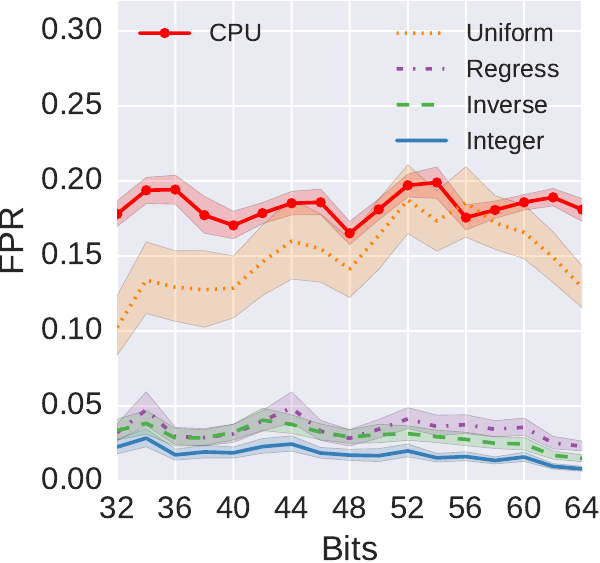}
\par\end{centering}
}
\par\end{centering}
\caption{Experimental results. Bands show 95\% confidence intervals obtained
over 100 different $n$ of equal magnitude.\label{fig:Experimental-results}}
\end{figure}

Figure \ref{fig:Experimental-results} summarizes the results. ROC
curves obtained using each quantization strategy for 64-bit $n$ are
shown in Figure \ref{fig:ROC-curve-summary}. The inverse strategy,
which has a 0.525\textpm 0.215\% equal error rate (EER), performs
nearly as well as the optimal integer strategy having a 0.318\textpm 0.220\%
EER. Results for the integer strategy are obtained by integrating
the bottom layer of the neuromorphic sieve off of the TrueNorth chip. 

Figure \ref{fig:Ticks-vs-bits} shows the number of clock cycles per
sieve value as a function of $\log_{2}n$ (bits). This metric remains
constant for the neuromorphic sieve, which performs a test for smoothness
within one clock cycle using any quantization strategy. CPU cycles
were measured using differences of the RDTSC instruction around sieving
routines on a single dedicated core of the host CPU. 

Figure \ref{fig:FPR-vs-bits} shows the FPR of each quantization strategy
at the point on the ROC curve where true positive rate (TPR) equals
the CPU TPR. For all $n$, and given the same level of sensitivity,
the neuromorphic sieve with binary weights (uniform strategy) demonstrates
equal or higher specificity, i.e., lower FPR, than the CPU-based sieve
and significantly lower FPR using quaternary weights (regress and
inverse strategies). The higher-precision integer weights performed
marginally better than quaternary weights.

\section{Conclusion \label{sec:Conclusion}}

This work highlights the ability of a neuromorphic system to perform
computations other than machine learning. A $O\left(1\right)$ test
for detecting smooth numbers with high probability is achieved, and
in some cases is significantly more accurate than a CPU-based implementation
which performs the same operation in $O\left(\ln\ln B\right)$ amortized
time. Despite this, the NS1e has a normal operating frequency of 1
KHz and wall clock time is only asymptotically lower than that of
the CPU. Future high-frequency neuromorphic architectures may be capable
of sieving large intervals in a much shorter amount of time. How such
topological and precision constraints determine the accuracy of these
architectures is an item for future work.

It is also worth noting that since the NS1e is a digital architecture,
at the hardware level it does not achieve constant time synaptic integration.
The ability to perform constant time addition is promised only by
analog architectures that exploit the underlying physics of the device
to compute, for example by integrating electrical potential (memristive
architecture) or light intensity (photonic architecture).

\bibliographystyle{plain}
\bibliography{neurosieve}

\begin{thebibliography}{1}

\bibitem{crandall2006prime}
Richard Crandall and Carl Pomerance.
\newblock {\em Prime numbers: a computational perspective}, volume 182.
\newblock Springer, 2006.

\bibitem{granville2008smooth}
Andrew Granville.
\newblock Smooth numbers: computational number theory and beyond.
\newblock {\em Algorithmic number theory: lattices, number fields, curves and
  cryptography}, 44:267--323, 2008.

\bibitem{merolla2014million}
Paul~A Merolla et~al.
\newblock A million spiking-neuron integrated circuit with a scalable
  communication network and interface.
\newblock {\em Science}, 345(6197):668--673, 2014.

\bibitem{montgomery1995block}
Peter~L Montgomery.
\newblock A block lanczos algorithm for finding dependencies over gf (2).
\newblock In {\em Advances in cryptology--EUROCRYPT'95}, pages 106--120.
  Springer, 1995.

\bibitem{pomerance1984quadratic}
Carl Pomerance.
\newblock The quadratic sieve factoring algorithm.
\newblock In {\em Advances in cryptology}, pages 169--182. Springer, 1984.

\bibitem{pomerance1994role}
Carl Pomerance.
\newblock The role of smooth numbers in number theoretic algorithms.
\newblock In {\em Proc. Internat. Congr. Math., Z{\"u}rich, Switzerland},
  volume~1, pages 411--422, 1994.

\end{thebibliography}

\end{document}